\documentclass[journal]{IEEEtran}
\ifCLASSINFOpdf
  \usepackage[pdftex]{graphicx}
\else
  \usepackage[dvips]{graphicx}
\fi
\usepackage{amsmath}
\usepackage{booktabs}
\usepackage{multirow}
\usepackage{tabularx}
\usepackage{booktabs}
\usepackage{lscape}
\usepackage{longtable}
\usepackage{lipsum}
\usepackage[dvipsnames]{xcolor}
\definecolor{myblue}{rgb}{0.36, 0.54, 0.66}
\definecolor{mygreen}{rgb}{0.53, 0.66, 0.42}
\definecolor{mypurple}{rgb}{0.6, 0.4, 0.8}
\definecolor{mypink}{RGB}{219, 112, 147}

\usepackage{algorithmic}
\usepackage{inconsolata}
\usepackage{tabularx}
\usepackage{graphicx}
\usepackage{subcaption}
\usepackage{url}

\usepackage{array}
\begin{document}
%
\title{A Gap in Time: The Challenge of Processing Heterogeneous IoT  Data in Digitalized Buildings}
%
%
%


\author{
    \IEEEauthorblockN{
    Xiachong Lin\IEEEauthorrefmark{1}\IEEEauthorrefmark{3}, 
    Arian Prabowo\IEEEauthorrefmark{1}\IEEEauthorrefmark{3}, 
    Imran Razzak\IEEEauthorrefmark{1}, 
    Hao Xue\IEEEauthorrefmark{1}, 
    Matthew Amos\IEEEauthorrefmark{2},
    Sam Behrens\IEEEauthorrefmark{2},
    Flora D. Salim\IEEEauthorrefmark{1}} 
    \IEEEauthorblockA{\IEEEauthorrefmark{1}University of New South Wales, Sydney, Australia} \\
    \IEEEauthorblockA{\IEEEauthorrefmark{2}CSIRO Energy Centre, Newcastle, Australia} \\
    Email: \{dawn.lin, arian.prabowo, imran.razzak, hao.xue1, flora.salim\}@unsw.edu.au \\
    \{matt.amos, sam.behrens\}@csiro.au \\
    \thanks{\IEEEauthorrefmark{3} Equivalent contribution.}
}

\maketitle

\begin{abstract}
The increasing demand for sustainable energy solutions has driven the integration of digitalized buildings into the power grid, leveraging Internet-of-Things (IoT) technologies to enhance energy efficiency and operational performance. Despite their potential, effectively utilizing IoT point data within deep-learning frameworks presents significant challenges, primarily due to its inherent heterogeneity. This study investigates the diverse dimensions of IoT data heterogeneity in both intra-building and inter-building contexts, examining their implications for predictive modeling. A benchmarking analysis of state-of-the-art time series models highlights their performance on this complex dataset. The results emphasize the critical need for multi-modal data integration, domain-informed modeling, and automated data engineering pipelines. Additionally, the study advocates for collaborative efforts to establish high-quality public datasets, which are essential for advancing intelligent and sustainable energy management systems in digitalized buildings.

\end{abstract}


%
\IEEEpeerreviewmaketitle
\begin{figure*}[!htb]
    \centering
    \includegraphics[width=0.95\textwidth]{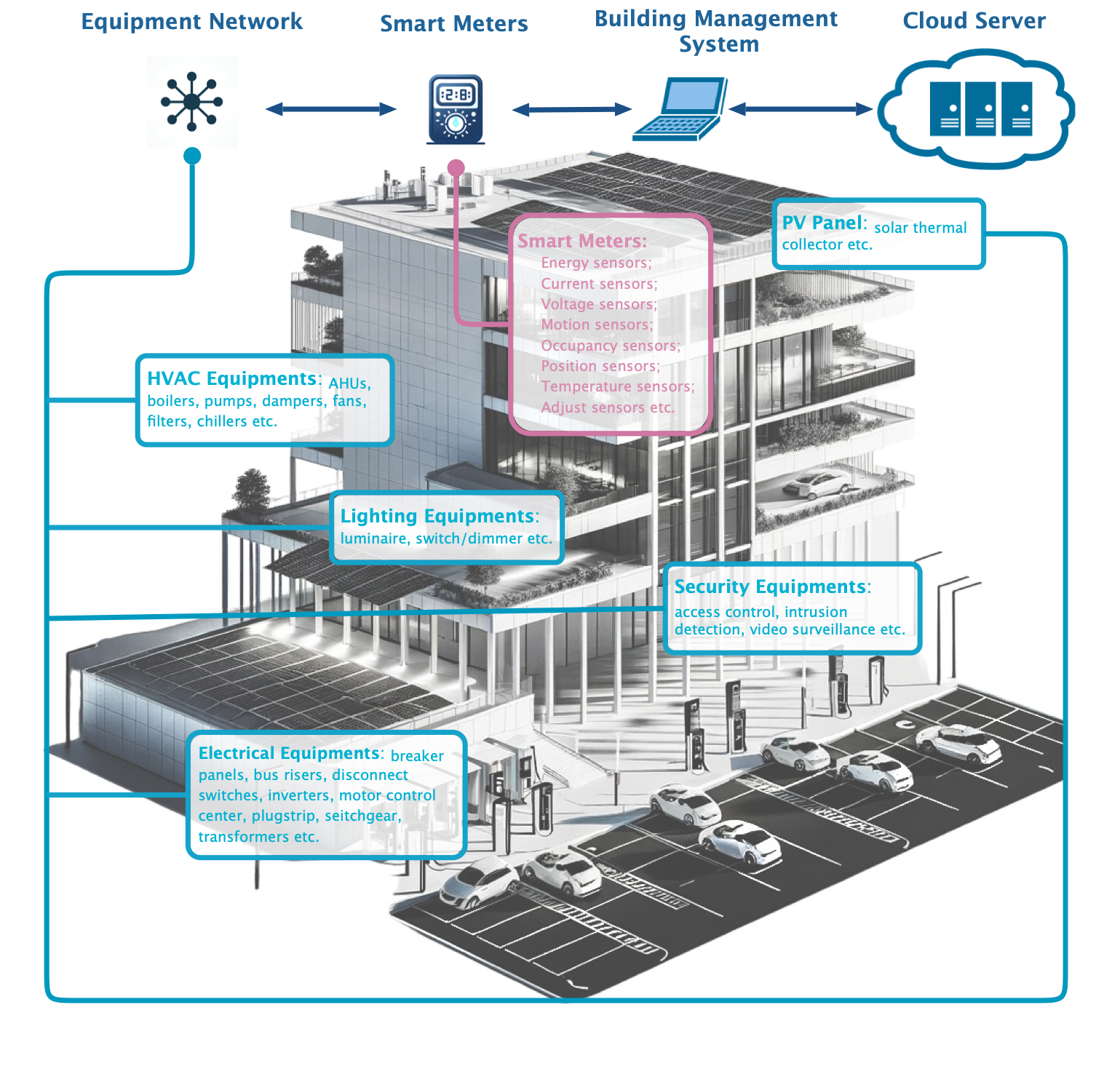}
    \caption{The schematic graphic about data communication in a digitalized building. The activities in a digitalized infrastructure are typically monitored by IoT sensors. Meter readings are uploaded to the Building Management System and then stored in the cloud-based server. This is a bidirectional conversation where the decisions made based on the data analytics results can intently impact the building operation. }
    \label{fig:building}
\end{figure*}
\section{Introduction}
\IEEEPARstart{T}he paradigm of energy production and consumption is shifting toward a model that prioritizes reliability, affordability, and environmental sustainability. This transformation, often referred to as the “4Ds” of energy—decarbonization, decentralization, digitalization, and democratization—is underpinned by a global consensus to reduce carbon emissions and foster energy independence\footnote{RACE for 2030 https://racefor2030.com.au/}. Nations are increasingly investing in renewable energy sources and innovative energy storage solutions to create resilient, digitally enhanced infrastructures that can respond to fluctuations in energy transmission and market volatility while complying with strict environmental regulations.

One area receiving significant attention within this energy transition is the integration of buildings into the energy grid. Buildings, as foundational components of urban environments, play a critical role in human well-being, impacting occupant comfort~\cite{gao2021transfer}, health, and safety~\cite{dong2021occupant}. They are also major consumers of global primary energy and substantial contributors to carbon dioxide emissions, rivaling the transportation sector~\cite{iea-ebc}. Improving building performance, therefore, holds immense potential for mitigating carbon emissions and addressing climate change. By sourcing renewable energy upstream from the grid while optimizing building operations downstream, energy savings can be realized, greenhouse gas emissions reduced, and indoor air quality improved. Moreover, resilient building infrastructures enhance adaptability in the face of climate-related challenges.

In addition to passive energy optimization, buildings have substantial potential to actively participate in energy systems through building-to-grid (B2G) interactions. In this context, buildings can serve as components of virtual power plants (VPPs), which are networks of decentralized power generation or storage units. Through B2G interactions, buildings can dynamically respond to grid signals by supplying excess onsite-generated energy or reducing demand during peak times. Compared to other bidirectional energy systems, such as vehicle-to-grid (V2G), B2G offers larger, more stable energy contributions without significantly impacting the user experience.

A Building Management System (BMS) is key to enabling these interactions, acting as a sophisticated network of hardware and software that intelligently controls and monitors building operations. As shown in Figure~\ref{fig:building}, a digitalized building equipped with an Internet of Things (IoT) network can continuously track activities related to HVAC systems, lighting, photovoltaic (PV) panels, security, and electrical facilities. The BMS collects data from IoT devices distributed throughout the building and uploads it to a cloud server, allowing for centralized and cloud-based management of IoT data streams for further analysis and optimization.

While advancements in IoT technology have led to an explosion in data generation within buildings, unlocking the full potential of this data for enhancing building efficiency and functionality faces several challenges. Privacy concerns, for instance, limit the availability of publicly accessible building data, reducing the data pool for cross-building analytics~\cite{jin2023review,jin2020accurate}. Additionally, the unique characteristics of each building make it difficult to deploy universal software solutions across diverse structures. IoT systems in buildings are typically implemented with customized schemas tailored to the specific requirements of each building or its management system, defining how data is organized, labeled, and accessed. This schema variability, influenced by differences in building types, purposes, and infrastructure, complicates data integration as buildings may use different terminologies, data formats, and organizational structures.

Addressing these issues in cross-building analytics requires Centralized Management Tools (CMT) to reconcile these differences. Initiatives like Project Haystack\footnote{Haystack https://project-haystack.org} and Brick Schema~\cite{balaji2018brick} have advanced efforts to standardize vocabularies and relationships for building systems, facilitating data integration and analysis across diverse buildings. However, despite these efforts, even though works as ~\cite{wang2020iot, amade2021identifying} discuss the IoT applicants in digitalized buildings for energy control, there is limited research discussing leveraging these standardized schemas within deep learning frameworks. This gap in the literature highlights an important area for future exploration and development, as integrating schema standardization with machine learning models could enhance the scalability and adaptability of smart building technologies.
This work specifically concentrates on utilizing real-world building IoT-generated data to illustrate the challenges inherent in building IoT analytics. It discusses the opportunities for future work and provides directions for bridging the gap between state-of-the-art techniques and real-world applications. 

The remainder of this paper is organized as follows: Section~\ref{sec:DataIntro} provides an overview of the building IoT data source and its components. Sections~\ref{sec:Intra}-\ref{sec:Inter} examine the challenges associated with processing IoT-generated data within individual buildings and across multiple buildings, respectively. A case study benchmarking state-of-the-art time-series forecasting methods on building IoT data is presented in Section~\ref{sec:Benchmark}. Section~\ref{sec:Discussion} summarizes the key challenges and opportunities in this domain, followed by the conclusion in Section~\ref{sec:conclusion}.

\section{Building IoT Data}
\label{sec:DataIntro}
This section provides an overview of how building IoT data, typically formatted by centralized management tools (CMTs), is structured. A building IoT dataset generally comprises the following components:

\subsection{Knowledge Graph}
In CMTs, the semantic relationships within a building are represented using a knowledge graph (KG). This graph encodes the spatial, functional, and operational aspects of the building, including location floor maps, equipment distribution, and IoT device installations. Generating such a KG for a digitalized building requires detailed information about the building's assets, spatial layout, and system hierarchies. Collaboration with building engineers or building information modeling (BIM) experts is often necessary to ensure accurate mapping of physical and functional relationships between components.

Using the \texttt{Brick Schema} as an example, nodes in the knowledge graph can represent the following categories:
\begin{itemize}
    \item \texttt{Location}: Specifies spatial areas within the building, such as \texttt{Floor, Conference Room, Hallway}.
    \item \texttt{Equipment}: Systems or mechanisms that serve the building, such as \texttt{Chiller, Boiler, Fan Coil Unit, Chilled Water System} and \texttt{Hot Water Pump}.
    \item \texttt{Point}: Refers to IoT devices installed to monitor building activities. A \texttt{Point} may fall into groups such as \texttt{Alarm}, \texttt{Parameter}, \texttt{Limit}, \texttt{Status}, \texttt{Setpoint}, or \texttt{Command}. By default, it is categorized as \texttt{Point}. Examples include \texttt{Current Sensor}, \texttt{Voltage Sensor}, \texttt{Cooling Demand Setpoint}, and \texttt{Emergency Alarm}.
\end{itemize}

The KG consists of a large number of triples, where each triple $(s, p, o)$ describes a relationship (predicate) $p$ between a subject node $s$ and an object node $o$.

\subsection{Data Stream}
Each \texttt{Point} or IoT device is associated with a unique stream identifier, linking it to a data stream stored in the server database. A data stream is a time series consisting of a collection of data points. Each data point $x_t$ represents the value of a meter reading at a specific timestamp $t$. These time series log building sensor readings over time are collected in various sampling frequencies, enabling continuous monitoring and analysis of building activities.

\subsection{Public Accessible Dataset}
\textbf{BTS}~\cite{prabowo2024bts}: A dataset covers three Australian buildings over three years, comprising more than ten thousand time series data points with hundreds of unique IoT categories. BTS-B is currently available for public access.

\textbf{BLDG59}~\cite{luo2022three} A three-year dataset supporting research on building energy control and analytics, providing the building IoT data from an office building located in California, USA.
\section{Challenges for Intra-Building Scenario}
\label{sec:Intra}
\subsection{Irregular Granularity}


Building IoT devices often exhibit significant variability in data resolution, update frequency, and communication protocols, depending on sensor functionality and operational requirements. Some sensors are configured to capture high-resolution, high-frequency data, while others provide sparse updates based on the needs of specific applications. In the BTS-B building, IoT point data is captured at various temporal granularities, including 1-minute, 5-minute, 10-minute, and daily intervals. Notably, even sensors of the same type can operate at different sampling frequencies, such as \texttt{Voltage Sensors} with some capturing data at 1-minute intervals and others at 5-minute intervals, reflecting the customized configurations for different use cases.

These variations in data frequency present challenges in synchronizing data across different streams, particularly when aggregating or analyzing data from multiple devices. High-frequency data streams may contain gaps due to network latency or sensor downtime, resulting in missing data points that must be addressed through interpolation or other gap-filling techniques to maintain data continuity. Conversely, lower-frequency data streams often need to be upsampled for consistency with higher-frequency data, a process that can introduce noise or artifacts if not handled carefully.

Addressing these challenges requires the implementation of advanced time-series analysis techniques capable of managing diverse temporal resolutions. Effective methods for resampling, interpolation, and noise reduction are essential to integrate data accurately across varying frequencies, ensuring the integrity and reliability of subsequent analysis. Developing robust data handling pipelines that account for the variability in IoT data sampling is critical to achieving accurate insights and supporting the broader objectives of smart building management and optimization.

\subsection{Long Tailed Distribution in IoT Device Category}
\begin{figure}
    \centering
    \includegraphics[width=\linewidth]{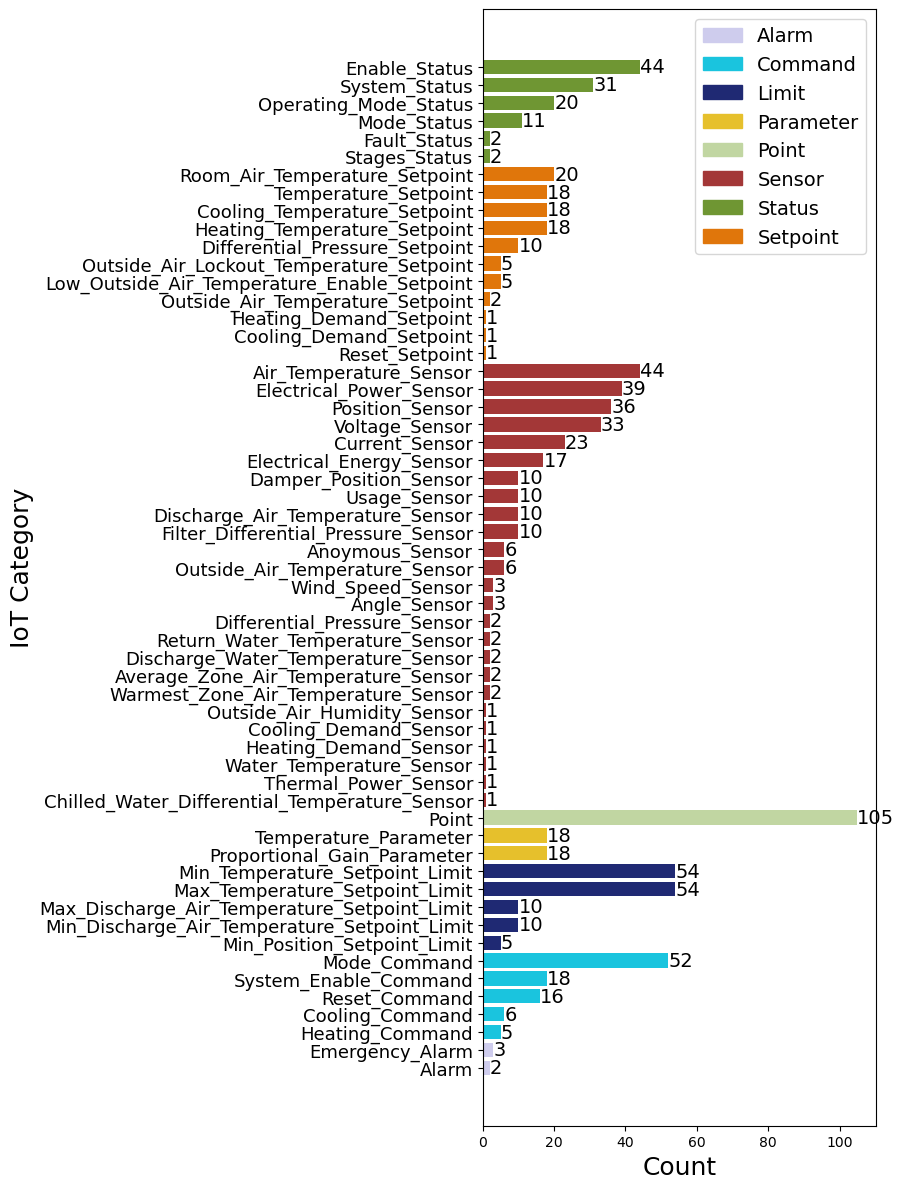}
    \caption{The statistics about the count number for each unique IoT category based on Brick ontology definition. All IoT categories are grouped into Alarm, Parameter, Limit, Status, Setpoint, Point, Command, and Sensor based on the device functionalities. The assignment of each unique IoT category to the corresponding group are represented by multiple colors as legend.}
    \label{fig:ontcount}
\end{figure}

Building IoT systems offer high-resolution and comprehensive insights into building activities and operations, characterized by a broader and more diverse range of sensor categories. Unlike typical time series datasets, such as electricity-focused datasets like \texttt{ETT}\cite{zhou2021informer} and \texttt{UCI Electricity}\cite{trindade2015UCIdataset}, which maintain feature category uniformity and primarily focus on specific sensor types like transformer temperature and electricity load~\cite{prabowo2023energyBuildSys,prabowo2023continually}, building IoT data encompass a wide array of sensor categories beyond these conventional measurements. This diversity adds significant complexity to data analysis and modeling.

In addition, the long-tailed distribution of IoT sensor categories results in substantial data imbalance, where certain sensor types have a large number of instances while others are underrepresented. For example, in the BTS-B building, sensors like \texttt{Enable Status} and \texttt{Air Temperature Sensor} have dense instances, whereas sensors such as \texttt{Reset Setpoint} and \texttt{Thermal Power Sensor} are sparsely deployed, as illustrated in Figure~\ref{fig:ontcount}. This distribution reflects the operational heterogeneity of building management systems, where sensor deployment is driven by specific functional requirements, leading to varying levels of representation across different sensor types.

The implications of this long-tailed distribution are significant. The data imbalance can skew analytical results, as models may be biased towards the more dominant sensor categories, potentially compromising the generalizability and accuracy of insights if this imbalance is not properly addressed. Furthermore, underrepresented categories pose additional challenges in training and validating machine learning models. Addressing this imbalance often requires specialized data preprocessing techniques or algorithmic adjustments to ensure that models can effectively learn from both well-represented and sparse sensor categories. These approaches are essential for achieving reliable and unbiased analyses in IoT-based building management systems.

\subsection{Heterogeneous Distribution in IoT-generated Data Streams}
\begin{figure*}[!htb]
    \centering
    \includegraphics[width=0.95\linewidth]{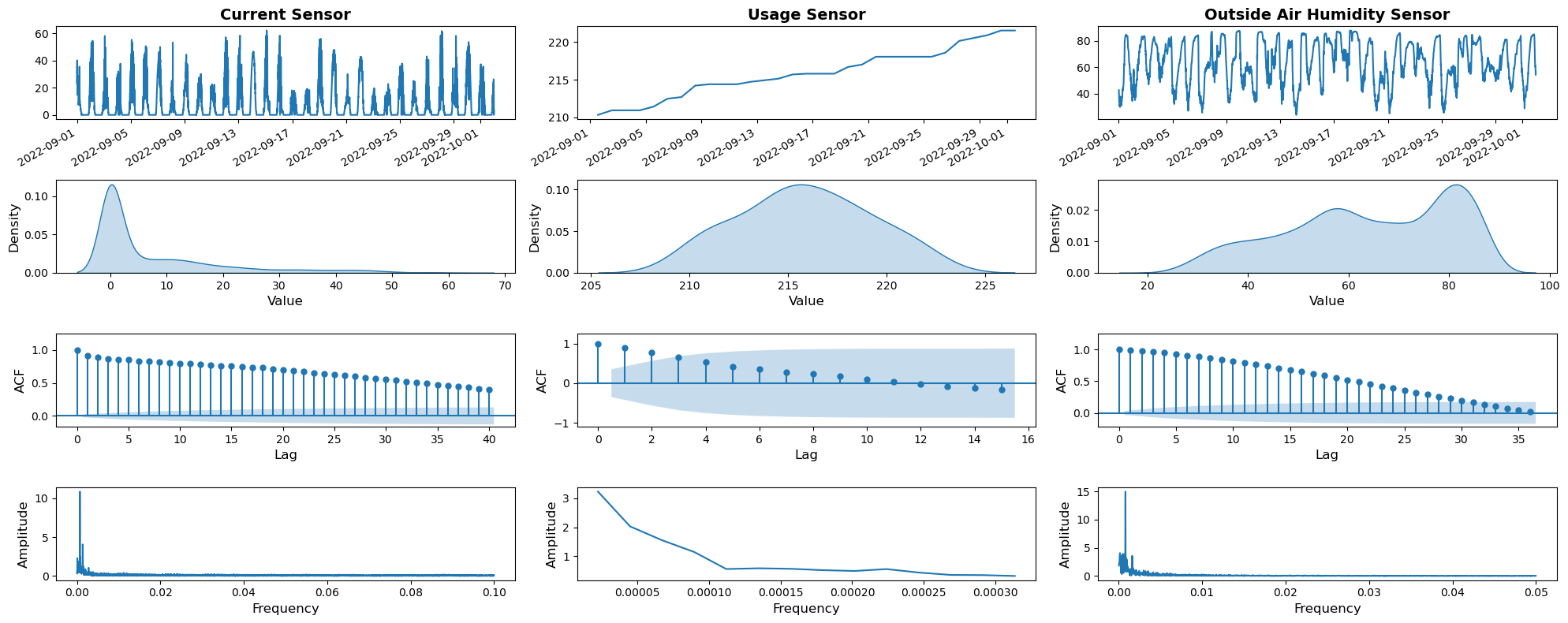}
    \caption{Distribution analysis for data stream segments from \texttt{Current Sensor, Usage Sensor} and \texttt{Outside Air Humidity Sensor}. Each row indicates the original data stream, Kernel Density Estimator fittings, Autocorrelation Function measuring, and Fourier Transformation from top to bottom. }
    \label{fig:datadist}
\end{figure*}

Heterogeneous distribution in IoT-generated data streams refers to the variability and diversity in the statistical properties of data collected from different IoT devices or even the same sensor under varying conditions. This heterogeneity arises from the dynamic and non-uniform behaviors of the monitored systems or environments, influenced by operational states, environmental factors, and the nature of the observed processes. Consequently, IoT data often exhibit non-stationary distributions, which can be highly skewed, bimodal, or multimodal. For instance, sensors may record bursts of activity interspersed with inactivity (e.g., energy usage spikes), continuous trends (e.g., cumulative usage), or alternating regimes (e.g., diurnal cycles in temperature or humidity). These characteristics introduce analytical challenges and necessitate tailored modeling approaches.

Figure~\ref{fig:datadist} illustrates the heterogeneity in one-month data segments collected from three IoT sensors. The \texttt{Current Sensor} demonstrates significant heterogeneity, with a highly skewed distribution dominated by values near zero and occasional extreme spikes. This pattern reflects a system with prolonged inactivity interrupted by sudden bursts of activity. In contrast, the \texttt{Usage Sensor} exhibits a smooth, unimodal distribution with a narrow spread, indicating a more homogeneous dataset that captures a steadily evolving process, such as cumulative usage. Meanwhile, the \texttt{Outside Air Humidity Sensor} displays moderate heterogeneity, characterized by a bimodal distribution with distinct peaks around 60 and 80. This likely represents alternating environmental regimes, such as day and night humidity levels.

A cross-sensor comparison highlights stark differences in the behaviors of the sensors. The \texttt{Current Sensor} and \texttt{Outside Air Humidity Sensor} exhibit high variability and periodicity, while the \texttt{Usage Sensor} is more stable and trend-oriented. These differences underscore the need for specialized modeling strategies: extreme value models for the \texttt{Current Sensor} to handle rare but significant spikes, trend-based models for the \texttt{Usage Sensor} to account for its gradual evolution, and state-dependent or bimodal models for the \texttt{Outside Air Humidity Sensor} to capture its alternating regimes.

Understanding and addressing such heterogeneity is critical for accurate modeling and analysis of IoT data. It necessitates the use of tailored techniques, such as feature scaling, clustering, or the application of specialized models capable of capturing periodicity, trends, and randomness inherent in the data. Effectively managing these diverse distributions not only enhances anomaly detection and forecasting accuracy but also facilitates the optimization of IoT systems, enabling more reliable and efficient operations across various applications.


\subsection{Complex Dependencies in IoT-generated Data Streams}
\begin{figure*}[htb]
    \centering
    \includegraphics[width=\linewidth]{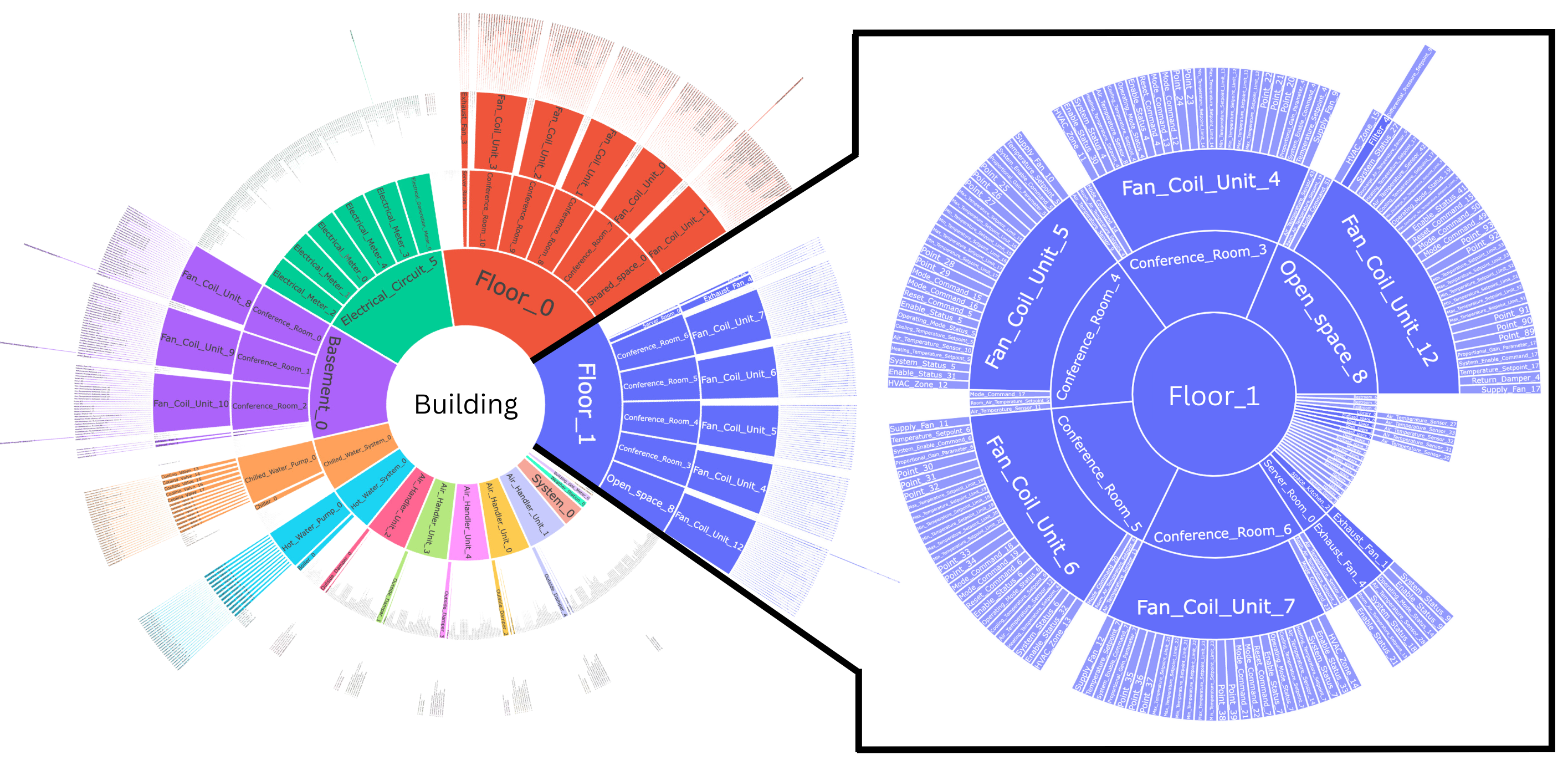}
    \caption{Visualization of hierarchical dependencies in the BTS-B building model. The inner circle represents the root of the hierarchy, with each successive outer layer iteratively mapping the top-down path from the root \texttt{Building} ontology through \texttt{Location} components, distributed \texttt{Equipment}, and finally, monitored \texttt{Point} sensors. The left side illustrates the overall hierarchical structure of the building, while the right side zooms in on a specific component \texttt{Floor}. This model structure is customized and varied by the unique design and IoT distribution in each registered building. }
    \label{fig:hierarchyComplex}
\end{figure*}

IoT-generated data in buildings offers extensive insights into various operational aspects, including energy consumption, environmental comfort, and HVAC (Heating, Ventilation, and Air Conditioning) performance. Despite the potential of this data, leveraging it effectively poses significant challenges due to the complexity introduced by the large number of IoT devices and the dependencies between sensors. The dependencies in IoT sensor data arise primarily from two key factors:

\begin{itemize}
    \item\textit{Physical Distribution of Sensors:} The spatial arrangement of sensors within a building—determined by their placement, proximity, and interaction with the environment—plays a critical role in shaping sensor dependencies. For instance, sensors located in close proximity may exhibit correlated behavior due to shared exposure to localized environmental conditions. Understanding these physical relationships is essential for accurately modeling sensor interactions and deriving meaningful insights.
    \item\textit{Schema Hierarchy in Centralized Management Tools:} The logical organization of sensors within centralized management tools defines the schema hierarchy, which dictates how sensors are grouped, categorized, and interconnected. This hierarchy affects the interpretation and flow of data across different system levels, influencing dependencies beyond the physical domain. Properly accounting for these logical relationships is crucial for effective data integration and analysis.
\end{itemize}

Figure~\ref{fig:hierarchyComplex} provides a real-world example about the complex hierarchical relations in BTS-B building. While the KG provide rich metadata, transforming this information into an analytics-ready format often requires substantial preprocessing and engineering effort.

The version updates of centralized management tool can introduce additional challenges as resulting in variations in data structures, sensor hierarchies, and feature definitions across framework versions. These inconsistencies can disrupt the deployment of pre-trained models in new buildings, necessitating significant engineering effort to adapt models to changes in schema and semantics.

\section{Challenges in Inter-Building Scenario}
\label{sec:Inter}
Inter-building analytics entails the examination of data from multiple buildings to gain insights and optimize building performance on a broader scale. Often applied in community or district-level building management, this approach leverages shared insights to drive efficiency improvements across a network of buildings. This technique is especially advantageous when data from individual buildings is limited, as aggregating information across multiple buildings can uncover patterns not visible in isolated datasets. Furthermore, cross-building analytics supports the creation of models that are generalizable and can be deployed across diverse buildings, enhancing scalability and adaptability. However, this approach also presents challenges in processing cross-building IoT data, including variations in spatial constraints and building structures, as well as schema variability and customization, which complicate data integration and model consistency across heterogeneous building environments.

\subsection{Spatial and Structural Differences}

In cross-building scenarios, one of the foremost challenges arises from the inherent spatial and structural differences among buildings. As illustrated in Figure~\ref{fig:hierarchyComplex}, the hierarchical structure of a building varies significantly depending on its design elements, including layout, size, floormap, and functional spaces. These factors directly influence the distribution and placement of IoT sensors and devices, creating unique spatial and operational contexts for each building. Such variations pose significant obstacles to aligning sensor data and interpreting spatial dependencies, as the physical characteristics of a building shape not only how data is collected but also how building systems interact and operate.

For instance, an open-floor office building, characterized by fewer partitions and large interconnected spaces, may exhibit distinct airflow dynamics and require a unique sensor placement strategy to monitor temperature, air quality, or occupancy. In contrast, a segmented office building with enclosed rooms may demand a denser and more localized sensor network to capture the diverse environmental conditions across its spaces. Similarly, older buildings, often retrofitted with IoT systems, may face architectural constraints such as limited access to wiring pathways or structural interference, making sensor placement and connectivity more challenging. Conversely, modern smart buildings are often designed with IoT integration in mind, featuring optimized sensor placement and pre-built infrastructure to support interconnected systems.

These spatial and structural differences impact not only the placement and functionality of IoT devices but also the ability to generalize models and algorithms across buildings. A model trained on data from one building may fail to capture the unique spatial dependencies and operational characteristics of another due to differences in layout or sensor distribution. Additionally, inconsistencies in sensor density, proximity, and coverage across buildings can lead to biased or incomplete interpretations of sensor data.

To address these challenges, flexible and adaptive data processing methods are essential. These methods must account for diverse building layouts, variations in sensor placement, and building-specific operational characteristics. For example, spatial alignment techniques can be used to map sensor data to a common reference framework, enabling cross-building analysis despite structural differences. Similarly, adaptive algorithms capable of dynamically adjusting to building-specific contexts can improve the robustness and scalability of IoT analytics.

\subsection{Inconsistent Data Representation}

A major challenge in cross-building analytics is the variability and heterogeneity in data representation, even when buildings utilize the same centralized management tools. While such tools aim to provide standardized frameworks for organizing and managing building IoT data, significant differences arise due to variations in building types, engineering objectives, and existing infrastructure. These differences lead to customized schemas that dictate how data is labeled, organized, and accessed within the system. 
For instance, building-specific customizations often include unique prefixes, tailored nomenclatures, or distinct hierarchical structures within the knowledge graph. Such customizations reflect localized requirements, such as specific equipment configurations, operational goals, or compliance with regional standards. However, these modifications create inconsistencies in the underlying data schema across buildings. Consequently, even when two buildings rely on the same CMT, their KGs may differ in structure, naming conventions, and metadata organization.

This heterogeneity complicates the task of developing generalizable solutions for IoT-enabled analytics. Extracting meaningful information from the KG requires designing pre-processing pipelines tailored to the specific schema of each building. For example, customized mechanisms may be needed to resolve naming conflicts, map unique prefixes to standardized terms, or interpret location and device metadata in a consistent manner. Without such tailored pre-processing, data inconsistencies can hinder effective cross-building comparison, integration, and analysis.

Addressing these challenges demands robust strategies that balance standardization with flexibility. On one hand, standardized data representation practices can help reduce variability by establishing common guidelines for schema design, labeling conventions, and metadata organization. On the other hand, adaptive processing techniques must be developed to handle building-specific customizations dynamically. By combining these approaches, it becomes possible to achieve scalable and interoperable IoT analytics across diverse building environments.

\section{Case Study: Benchmarking on Building IoT Forecasting}
\label{sec:Benchmark}
To assess the adaptability of deep-learning frameworks to building IoT data, this section presents a benchmark study that evaluates the performance of state-of-the-art algorithms on the BTS-B building IoT datasets.

\subsection{Problem Formulation}

Denote the dataset which contains $K$ time steps and $N$ IoT points as $X \in \mathcal{R}^{N \times K}$. Each training sampling is denoted as $d_{N, k} = D_{N, k:k+S}$, where $S$ is the sequence length of the historical data. Denote the forecasting model as $f(\cdot)$. The multi-variate multi-step forecasting is formalized as $f(d_{N, k}) = d_{N, k+S+H}$, where $H$ is the forecasting horizon. In this setting, $S=12$, $H=12$.

\subsection{Dataset and Pre-processing}
This study utilizes BTS-B~\cite{prabowo2024bts} as the dataset. This dataset contains 731 non-zero IoT data streams collected from an anonymous building in Australia. To construct the regular training data, we slice the data by date from 00:00:00 01/05/2022 to 00:00:00 01/10/2022 and set the granularity to 10 min. Missing values caused by temporal irregularity are filled with the mean value. To simplify the modeling process, we drop point data whose missing ratio \texttt{r\_missing > 2\%} and \texttt{n\_Unique > 1,000}. A Standard Normalization process is conducted for scaling. Consequently, we obtained the 5-month training data containing 22,033 time steps and 166 distinct IoT point data. Sequentially, we split the data into training, validation, and testing, using the ratio of 0.7, 0.1, 0.2.

\subsection{Evaluation Metrics}
MSE, RMSE, MSE, SMAPE, and R$^2$ are employed for model performance evaluation. Each metric score is computed separately for each forecasting step, followed by the average.

\subsection{Experimental Settings}
We employ the Adam optimizer with a 0.001 learning rate and the MSE loss to guide the multivariate forecasting and transfer learning experiments. Training epochs are default to 10 with an EarlyStopper, which \texttt{patience=3}. Besides, a LearningRateScheduler with \texttt{factor=3} reduces the learning rate by half at every epoch. All experiments are completed on an A100 GPU acceleration.

\subsection{Baseline Models}
We selected seven state-of-the-art time series models as baseline comparisons, covering diverse architectural paradigms to ensure a comprehensive evaluation. These include graph-based models such as GWaveNet \cite{wu2019graph} and SCPT \cite{prabowo2023traffic}; transformer-based models such as Informer \cite{zhou2021informer}, iTransformer \cite{liu2023itransformer}, and PatchTST \cite{nie2022time}; an MLP-based model, DLinear \cite{zeng2023transformers}; and a convolutional model, TimesNet \cite{wu2022timesnet}.

Among these, DLinear and PatchTST are Channel-Independent (CI) models, which process each IoT data stream independently without explicitly modeling interactions between channels. In contrast, the graph-based models (GWaveNet and SCPT), convolutional model (TimesNet), and transformer-based models (Informer and iTransformer) are Channel-Dependent (CD) models. These CD models aim to capture dependencies and interactions across channels, which can be beneficial for learning complex relationships but may also face challenges when handling heterogeneous IoT data streams, particularly in scenarios with a high number of channels or significant variations across streams.

\subsection{Result Analysis}
 \begin{table}[!htb]
    \centering
    \begin{tabular}{@{}ll|ccccc@{}}
    \toprule
        \multicolumn{2}{c}{} & MAE & RMSE & MSE & SMAPE & R$^2$ \\
             \midrule
         \multirow{5}{*}{\rotatebox[origin=c]{90}{GWaveNet}}  
            & mean & 0.3433   &   0.7451   & 0.5594  &  25.8277 & 0.4320 \\
            & std  & 0.0694 & 0.0978 & 0.0676 & 6.4912 & 0.1034 \\
            & 25\%  & 0.2998 & 0.4915 & 0.7010 & 21.9758 & 0.3585 \\
            & 50\%  & 0.3637 & 0.5874 & 0.7664 & 27.3963 & 0.4044 \\
            & 75\%  & 0.3924 & 0.6266 & 0.7916 & 30.5799 & 0.5026 \\
            \midrule
         \multirow{5}{*}{\rotatebox[origin=c]{90}{Informer}}  
            & mean & 0.3048	& 0.7378   &  0.5463   & 21.7450 & 0.4004  \\
            & std &  0.0390 & \textbf{0.0684} & 0.0467 & 3.0093 & 0.0942 \\
            & 25\%  & 0.2774 & 0.4968 & 0.7048 & 19.6322 & 0.3283 \\
            & 50\%   & 0.3083  & 0.5506  & 0.7420  & 21.9284  & 0.4012 \\
            & 75\% & 0.3351 & 0.5984 & 0.7736 & 24.1005 & 0.4704 \\
            \midrule
        \multirow{5}{*}{\rotatebox[origin=c]{90}{DLinear}}  
            & mean & 0.3276  &  0.7185    &  0.5196   &  21.2940 & \textbf{0.4523} \\
            & std &  0.0694 & 0.0860 & 0.0606 & 4.7086 & 0.0931 \\
            & 25\% &  0.2826  & 0.4592 & 0.6776 & 18.2835 & \textbf{0.3815} \\
            & 50\%  & 0.3373 & 0.5266 & 0.7256 & 21.9352 & \textbf{0.4437} \\
            & 75\%  & 0.3807 & 0.5844 & 0.7645 & 24.8691 & \textbf{0.5174} \\
            \midrule
        \multirow{5}{*}{\rotatebox[origin=c]{90}{SCPT}}  
            & mean &  0.3399 &   0.7347   & 0.5438    &  24.7975 & 0.4481 \\
            & std &  0.0575 & 0.0951 & 0.0659 & 4.4296 & 0.0986 \\
            & 25\%  & 0.3027 & 0.4771 & 0.6907 & 21.8089 & 0.3681 \\
            & 50\% &  0.3501 & 0.5544 & 0.7446 & 25.4500 & 0.4363 \\
            & 75\% &  0.3866 & 0.6217 & 0.7884 & 28.1899 & 0.5160 \\
            \midrule
        \multirow{5}{*}{\rotatebox[origin=c]{90}{TimesNet}}  
            & mean & 0.3075   &  0.7458    & 0.5581 &  21.4314   & 0.4055   \\
            & std  & \textbf{0.0385} & 0.0685 & \textbf{0.0461} & \textbf{2.9658} & \textbf{0.0901} \\
            & 25\%  & 0.2804 & 0.5082 & 0.7129 & 19.3511 & 0.3367 \\
            & 50\%  & 0.3106 & 0.5601 & 0.7484 & 21.7264 & 0.4034 \\ 
            & 75\%  & 0.3375 & 0.6110 & 0.7817 & 23.7232 & 0.4727 \\
            \midrule
        \multirow{5}{*}{\rotatebox[origin=c]{90}{PatchTST}}  
            & mean & \textbf{0.2739} & \textbf{0.7105} & \textbf{0.5077} & \textbf{19.4184} & 0.4381 \\
            & std & 0.0492 & 0.0778 & 0.0555 & 3.7578 & 0.1049 \\
            & 25\% & \textbf{0.2427} & \textbf{0.4545} & \textbf{0.6742} & \textbf{17.0819} & 0.3594 \\
            & 50\% & \textbf{0.2818} & \textbf{0.5157} & \textbf{0.7181} & \textbf{20.0105} & 0.4311 \\
            & 75\% & \textbf{0.3102} & \textbf{0.5655} & \textbf{0.7520} & \textbf{22.2231} & 0.5119 \\
            \midrule
        \multirow{5}{*}{\rotatebox[origin=c]{90}{iTransformer}}  
            & mean & 0.3047	& 0.7379  &  0.5466   &  21.6993 & 0.4028 \\
            & std 	& 0.0389	& 0.0686	& 0.0468	& 2.9710	& 0.0919 \\
            & 25\% 	& 0.2774	& 0.4970	& 0.7049	& 19.6277	& 0.3322 \\
            & 50\% 	& 0.3084	& 0.5500	& 0.7416	& 21.9680 & 0.3996 \\
            & 75\% 	& 0.3350	& 0.5988	& 0.7738	& 24.0214	& 0.4705 \\
        \bottomrule
    \end{tabular}
    \caption{Empirical results for multi-step forecasting on BTS-B IoT data streams. The best score for each metric is highlighted in \textbf{Bold}. }
    \label{tab:demoexp_res}
\end{table}

As shown in Table \ref{tab:demoexp_res}, PatchTST consistently achieves superior performance across all evaluated metrics, recording the lowest mean errors for MAE, RMSE, MSE, and SMAPE at the 25\%, 50\%, and 75\% quantiles. These results highlight its accuracy and reliability in predicting time-series values relative to ground truth. For the average 12-step prediction horizon, PatchTST demonstrates notable improvements over GWaveNet, with reductions of 20.20\%, 9.25\%, 4.64\%, and 24.82\% across MAE, RMSE, MSE, and SMAPE, respectively. Similarly, it outperforms Informer by 10.12\%, 7.07\%, 3.69\%, and 10.70\% for the same metrics. These consistent gains underline the effectiveness of PatchTST in addressing the complexities of building IoT forecasting. However, the prediction errors remain significantly higher compared to benchmark datasets, reflecting the inherent challenges posed by heterogeneous and irregular IoT data streams.

One notable finding is the superior performance of channel-independent models, such as PatchTST and DLinear, when applied to the heterogeneous building IoT dataset. This suggests that CI models are better equipped to handle decentralized and irregular data characteristics, which are common in large-scale building systems. In contrast, channel-dependent models, such as Informer and iTransformer, often face difficulties in learning dependencies when the number of channels is large. Despite their advantages, CI models exhibit a key limitation: the inability to capture cross-channel dependencies and interactions. This shortfall limits their ability to leverage contextual and domain-specific relationships, which are crucial for building IoT systems with interconnected components.

Furthermore, the results underscore the broader limitations of the evaluated models in integrating domain knowledge and leveraging spatial graph structures inherent to building systems. Current state-of-the-art models demonstrate an ability to capture some dependencies without the use of additional metadata; however, their performance is constrained by the single-modal paradigm. This limitation highlights a critical challenge in the field: deep learning models relying exclusively on single-modal IoT data streams cannot fully address the complexities of building operations. To address this, future research should prioritize multi-modal approaches, particularly the integration of large language models, to effectively incorporate metadata, spatial interdependencies, and domain-specific insights. Multi-modal frameworks are not merely complementary but essential to unlocking the potential for intelligent, adaptive, and efficient analytics in building IoT systems.

\section{Challenges and Opportunities}
\label{sec:Discussion}

The application of AI technologies in the building IoT domain offers transformative potential for data analytics and energy management. However, several critical challenges persist, creating a gap between cutting-edge AI technologies and their practical deployment in digitalized building systems. These challenges can be broadly categorized into areas as below:

\subsubsection{Temporal Irregularity in IoT Data} Building IoT data streams are characterized by temporal irregularities, including varying sampling rates, missing data points, and unpredictable fluctuations in sensor readings. Such irregularities complicate time-series modeling, as traditional algorithms often assume regular and complete data streams. Additionally, the hierarchical complexity of building infrastructures introduces challenges in aligning temporal data with spatial and contextual metadata. Effective solutions must incorporate robust data preprocessing, imputation techniques, and adaptive time-series models capable of handling these irregularities.

\subsubsection{Heterogeneity in Building Layouts and IoT Distribution} In cross-building scenarios, the heterogeneity in building layouts, functional spaces, and IoT sensor distributions poses a significant barrier to algorithm generalization. Each building presents unique characteristics, such as varying floor plans, equipment types, and sensor placement strategies. These differences limit the ability of models trained on one building's data to perform well in another, necessitating substantial customization and retraining. Future research should focus on developing generalized algorithms that enable seamless model migration and knowledge transfer across buildings, thereby enhancing the scalability and applicability of AI-driven solutions.

\subsubsection{Integration of Multi-modal Data} Building IoT systems generate diverse types of data, including time-series sensor readings, spatial metadata, and domain-specific knowledge encoded in regulatory constraints and operational hierarchies. Integrating these heterogeneous data modalities into a unified framework remains a significant challenge. Current approaches often fail to capture the contextual intricacies required for comprehensive energy management and operational optimization. A promising research direction is the adoption of multi-modal learning techniques that embed domain knowledge into model architectures, facilitating holistic and context-aware analytics.

\subsubsection{Lack of Comprehensive Public Datasets} High-quality, comprehensive building IoT datasets are scarce, limiting the research community's ability to develop and benchmark innovative algorithms. Publicly available datasets often lack the diversity and resolution needed to represent the complexities of real-world building environments. This underscores the need for collaborative efforts between academia, industry, and regulatory bodies to create and share datasets that capture the breadth of building IoT scenarios. These datasets should include well-annotated multi-modal data, encompassing time-series readings, spatial hierarchies, and domain-specific metadata.

\subsubsection{Data Engineering Requirements}
Raw IoT data from real-world buildings is often fragmented, noisy, and incompatible with standard analytical frameworks. Significant data engineering efforts are required to preprocess, clean, and standardize this data to make it analytics-friendly. This includes resolving inconsistencies in naming conventions, aligning disparate data streams, and normalizing metadata representations. Automating these preprocessing steps and designing analytics-ready data pipelines are critical for reducing the time and effort required to derive actionable insights from IoT systems.

Despite the challenges, the digitalized building domain offers significant opportunities for innovation. Addressing the existing challenges can unlock transformative advancements in this field. For instance, developing generalized AI models that can operate across heterogeneous building environments will enable scalable solutions for cross-building analytics and model migration. Incorporating domain knowledge and integrating diverse data modalities into multi-modal learning architectures can facilitate context-aware and actionable insights. Additionally, collaborative efforts to establish comprehensive, high-quality public datasets will accelerate innovation and standardization. Intelligent IoT-driven systems can further reduce energy consumption while maintaining or improving comfort and compliance with regulatory standards. Automating data preprocessing pipelines will also lower the barrier to entry, making AI solutions more accessible and practical for real-world deployment. By leveraging these opportunities through innovative designs and collaborative initiatives, the building IoT domain has the potential to revolutionize energy management, operational efficiency, and sustainability in modern buildings.

\section{Conclusion}
\label{sec:conclusion}

Digitalized buildings offer significant opportunities for energy optimization and operational efficiency, but effectively leveraging IoT-generated data and integrating AI technologies remains a challenge. This paper identified key issues, including temporal irregularity, spatial heterogeneity, and the lack of standardized datasets, and discussed potential solutions in both intra-building and inter-building scenarios. Additionally, a benchmarking study highlighted the capabilities and limitations of state-of-the-art algorithms on building IoT data streams. Future work should focus on developing scalable, generalizable frameworks and fostering the creation of comprehensive public datasets to advance intelligent and sustainable building management systems.

\section*{Acknowledgment}
This research is funded by the NSW Government through CSIRO’s NSW Digital Infrastructure Energy Flexibility (DIEF) project as part of the Net Zero Plan Stage 1: 2020-2030, and by the Reliable Affordable Clean Energy for 2030 (RACE for 2030) Cooperative Research Centre.

\ifCLASSOPTIONcaptionsoff
  \newpage
\fi

\bibliographystyle{IEEEtran}
\bibliography{biblo} 

\end{document}